# Psychological Imagination Networks Show Cross-Population Centrality and Clustering Alignment in Humans That Large Language Models Fail to Replicate

Saurabh Ranjan and Brian Odegaard

University of Florida

**Author Note**

*This manuscript is preprint and not peer reviewed.*

ORCID:

Saurabh Ranjan: https://orcid.org/0000-0002-7868-7223

Brian Odegaard: https://orcid.org/0000-0002-5459-1884

Address Correspondence to:

Saurabh Ranjan: ranjan.saurabh@outlook.com

Brian Odegaard: bodegaard@ufl.edu

Competing interests: No competing interests to declare.

Author Contributions:

S.R.: Conceptualization, Visualization, Software, Methodology, Formal analysis, Data curation, Writing – original draft, Writing – review & editing.

B.O.: Writing – review & editing, Supervision.

Funding: S.R. was supported by the Threadgill Dissertation Fellowship from the University of Florida for this work.

Code and Data Availability: Code and data will be made available upon publication

## Abstract

Mental imagery vividness is a stable individual trait, yet whether imagined scenarios share relational structure across human and synthetic large language model (LLM) populations remains unknown. We applied psychological network analysis to vividness ratings from two validated questionnaires: the Vividness of Visual Imagery Questionnaire (VVIQ-2) and the Plymouth Sensory Imagery Questionnaire (PSIQ), across geographically and linguistically distinct human samples (Florida, Poland, and London; total $N = 2{,}743$) and six large language models (LLMs; Gemma3-12B/27B, their quantization-aware counterparts, Llama3.3-70B, and Llama4-16x17B). Imagination networks were constructed as regularized partial correlation graphs, with node centrality and community structure compared across populations using Pearson correlations and the Adjusted Rand Index (ARI). Human networks showed robust cross-population centrality correlations for expected influence, strength, and closeness ($r$ = 0.31-0.93), and community detection recovered clusters aligned with VVIQ-2 scene contexts (ARI = 0.27-0.40) and PSIQ sensory modalities (ARI = 0.87–1.0). Betweenness centrality was unstable across all populations, consistent with its sensitivity to individual experiential history. LLMs failed to replicate human network structure: LLM–human centrality correlations were weak and largely non-significant after correction, and most LLM configurations produced degenerate single-cluster topologies (median ARI = 0). This failure was consistent across model architectures, parameter scales (12B–272B), and conversational conditions. We posit that these findings may be driven by human imagination networks reflecting memory organization accumulated through embodied experience, a representational structure that linguistic training alone does not reproduce regardless of model scale and conversational memory.

**Psychological Imagination Networks Show Cross-Population Centrality and Clustering Alignment in Humans That Large Language Models Fail to Replicate**

Research on mental imagery vividness has established that it is a reliable, stable trait (Clark & Maguire, 2020; Schwarzkopf et al., 2025), that imagery strength is implicated in a range of cognitive processes from working memory to episodic memory (Pearson, 2019; Spagna et al., 2024a, 2024b, Ranjan & Odegaard, 2024), and that latent factor structure replicates across samples (Andrade et al., 2014; Jankowska & Karwowski, 2023). What research to date has not examined is whether there is relational structure among imagined scenarios, whether the same scenarios are structurally central across independent samples, and how they cluster. Here, we address these questions using two of the most widely used vividness measures: the Vividness of Visual Imagery Questionnaire (VVIQ-2; Marks, 1995), which probes the clarity and detail of visual scenes across eight everyday contexts, and the Plymouth Sensory Imagery Questionnaire (PSIQ; Andrade et al., 2014), which assesses vividness across seven sensory modalities. We then test whether large language models (LLMs), which acquire knowledge of imagery from language rather than from experience, produce networks whose centrality structure and community organization match those of human populations.

The VVIQ-2 asks participants to imagine eight everyday scene contexts, such as a friend's face, a rising sun, a country scene, and a railway station, and to rate the vividness of four aspects within each of eight contexts, yielding 32 items in total. The PSIQ covers seven modalities (visual, auditory, olfactory, gustatory, tactile, bodily sensation, and emotional feeling), sampling a wider sensory range with fewer items per modality, and includes imagination scenarios such as touching fur or a soft towel, smelling burning wood or fresh paint, or feeling scared or relieved. Factor-analytic work on the VVIQ consistently recovers a strong general imagery factor alongside weaker context-specific factors corresponding to the questionnaire's scene categories (Campos, 2011; Jankowska & Karwowski, 2023). Confirmatory factor analysis of the PSIQ recovers seven factors corresponding to the seven sensory modalities rather than a single general imagery factor; this modality-specific structure has been replicated across different language translations (Andrade et al., 2014; Pérez-Fabello & Campos, 2020).

Factor models and total scores decompose or aggregate covariance, which means associations among items are treated as a reflection of a common cause rather than as a relational map from which structural properties like centrality and clustering can be read. Whether centrality (the structural importance of individual scenarios in a network) is consistent across populations, and whether the community structure of that network aligns across populations, cannot be answered once observed variables are collapsed onto a latent dimension (Borsboom & Cramer, 2013). Using network psychometrics (Borsboom et al., 2021; Borsboom & Cramer, 2013), one can map the partial correlation structure among items directly, leaving the relational architecture intact and available for analysis that is an alternative, data-based approach to latent factor modelling.

The theoretical motivation for expecting structured item-level covariation comes from what imagination requires of memory. The subjective quality of imagining a scenario depends on how richly the relevant content is encoded in stored knowledge (Hassabis & Maguire, 2009; Schacter & Addis, 2007b, 2007a), and memory is organized around environmental regularities and sensory modalities rather than stored as a flat archive (Barsalou, 2008). Scenarios grounded in overlapping representational substrates should co-vary in vividness across a population; scenarios drawing on distinct substrates should vary more independently. The scenes and sensory experiences probed by the VVIQ-2 and PSIQ reflect common human experience whose semantic structure is broadly shared across languages and populations (Liang et al., 2024), so the partial correlation structure of vividness ratings should replicate across geographically and linguistically distinct samples.

An imagination network, as we define it, is a graph in which each node is an imagined scenario and each edge is the regularized partial correlation between two scenarios' vividness ratings, estimated via the EBICglasso method (Epskamp et al., 2018; Epskamp & Fried, 2018). EBICglasso is a method that uses a regularized partial correlation algorithm combined with the Extended Bayesian Information Criterion (EBIC; Friedman et al., 2008; Foygel & Drton, 2010). The glasso algorithm estimates a sparse precision matrix by applying an L1-regularization penalty (least absolute shrinkage and selection operator; LASSO) to the partial correlation coefficients. This penalty shrinks small, spurious associations to exactly zero, minimizing the risk of false-positive edges and enhancing model interpretability.

After computing the network over different populations, we measure node centrality that characterizes the structural importance of each scenario within this graph. Expected influence and strength index local importance by summing signed and unsigned edge weights to immediate neighbors, making them sensitive to direct vividness coupling. Closeness indexes global integration via shortest-path efficiency across the whole network. Betweenness counts how often a node lies on the shortest path between other node pairs, identifying scenarios that serve as bridges between otherwise weakly connected regions. Alternatively, we also compute the number of modular community structures that can be identified by the walktrap algorithm (Pons & Latapy, 2005), and measure cross-network alignment of communities, quantified by the Adjusted Rand Index (ARI; Hubert & Arabie, 1985).

Evaluating centrality correlations and measuring clustering/community alignment (using ARI) generates two predictions for population-level network properties that the present data can test. First, centrality values should correlate across human samples if they share similar world models (Balaban, 2026; Balaban & Ullman, 2025; Barsalou, 2008; Land, 2014; Liang et al., 2024; Richens et al., 2025). This prediction should hold most strongly for expected influence and strength, which reflect direct vividness coupling with related scenarios. Both closeness and betweenness are global centrality measures that rely on shortest-path calculations and have been questioned in psychological network contexts on conceptual grounds (Bringmann et al., 2019). The two measures capture fundamentally different aspects of global structure. Closeness indexes overall network integration efficiency and may prove replicable across populations, as it reflects broadly shared memory architecture. Betweenness identifies bridging scenarios whose role is sensitive to individual differences in experiential history; high within-population variability in those individual differences undermines cross-sample replicability (Bringmann et al., 2019). Second, community detection should recover clusters that align with the scene contexts of the VVIQ-2 and the sensory modalities of the PSIQ. Given the distinct neurobiological foundations of sensory modality organization relative to episodic scene organization, PSIQ clusters are predicted to be more tightly conserved across populations than VVIQ-2 clusters.

Interpreting the vividness ratings from LLMs requires a conceptual distinction about kinds of vividness and human cognitive processing to be able to translate it linguistically as drawn by Langkau (2021). Vividness is thought to be driven by two components: accuracy, and intensity. Accuracy refers to how closely a mental image imitates perceptual content clarity, detail, brightness and explains the epistemic uses of imagination. Intensity refers to the overall phenomenological force of an experience, including its affective and sensorimotor engagement, without requiring perceptual accuracy. The two can dissociate: a careful imagining of a clock face may score high on accuracy and low on intensity; a scene from a novel may score inversely. The VVIQ-2 and PSIQ items blend both senses. LLMs trained on text have learned the linguistic vocabulary of both accuracy and intensity and can produce plausible item-level ratings, which is why their outputs are not random.

Emerging evidence with LLMs suggests that even minimal sensory-style instructions (e.g., “imagine seeing…”) can guide text-only LLMs to organize their internal representations in ways that parallel modality-specific encoders (Wang et al., 2025). Thus, prompting may be sufficient to evoke structured, perceptual-like processing (Kawakita et al., 2024; Marjieh, Sucholutsky, et al., 2024; Wang et al., 2025). However, we posit that what they lack is the experience-grounded structure that organizes vividness across scenarios: for humans, scenarios sharing an environmental context cluster in vividness because they draw on overlapping episodic and semantic memory traces, not because they are linguistically similar. An LLM that assigns each item a plausible rating without cross-item memory structure produces a flat network with no coherent clusters and no conserved centrality. The network comparison in this research therefore probes not whether LLMs can produce plausible vividness language but whether their ratings carry the relational structure that, in human populations, traces memory organization.

In our work, six LLM variants were included: Gemma3-12B, Gemma3-27B, and their quantization-aware trained counterparts, along with Llama3.3-70B and Llama4-16x17B (a mixture-of-experts architecture with total 272B parameters). Each model completed the VVIQ-2 and PSIQ under two conversational conditions. In the independent condition, each item was rated without access to prior responses, matching the instructions given to human participants. In the cumulative condition, conversational history was maintained across items, giving models a form of within-session consistency analogous to the episodic memory that human participants naturally draw upon. The range of architectures, scales (12B-272B parameters), and conditions was chosen to test whether any currently available configuration approximates human imagination network structure.

Human imagination networks showed consistent centrality alignment and scenario clustering across independent population samples from the United States, Poland, and London, and this structure replicated across both the VVIQ-2 and the PSIQ, from visual scene imagery to seven-modality sensory imagination, respectively. LLM networks showed neither: centrality correlations with human networks were weak and largely non-significant after correction for multiple comparisons, most configurations produced degenerate single-cluster topologies, and the gap persisted across all model variants and both conversational conditions. This lack of similarity between human and LLMs is contrary to LLMs showing approximate human perceptual similarity judgments for color and other sensory dimensions, a result consistent with a qualitative difference in representational organization rather than a difference of degree (Kawakita et al., 2023; Marjieh et al., 2023; Marjieh, Kumar, et al., 2024; Wang et al., 2025). In the Discussion, we interpret these results through Balaban and Ullman's (2025) physics-graphics framework of mental simulation, examine two accounts of LLM failure alongside possible alternate explanations for the closeness centrality exception, address limitations including sample composition and the absence of multimodal models, and consider what the findings mean for cognitive science and for benchmarking generative AI world models.

## Method

### Participants

There were 541 participants (394 females and 147 males) in the VVIQ-2 Florida data, with a mean age of 20.4 years. The Florida dataset for both the VVIQ-2 and PSIQ was curated from responses provided by participants at the University of Florida, who received course credit for completing the task successfully. The data was collected from participants who gave consent prior to the start of the study. The study was approved by the Institutional Review Board of the University of Florida (IRB No.: ET00045024). For the Polish VVIQ-2 data, 1651 participants (1043 females and 608 males) fully completed the questionnaire from Jankowska & Karwowski, 2023, which comprises individuals from schools in Poland at the undergraduate level of college available here: https://osf.io/vh4es/. We created two non-overlapping subsets of 600 participants by randomly sampling the Polish dataset as *"Polish 1"* and *"Polish 2,"* whereas *"Polish All"* consists of all the data (1651 participants) from Poland in the results. We also analyzed the data by pooling all the data from the "Florida + Poland" group, which consisted of 2192 individuals.

For the PSIQ, there were 334 participants (276 females and 58 males) from Florida with a mean age of 20.5 years, and these participants also performed the VVIQ-2. There were 217 participants in the London dataset, which from Clark & Maguire (2023), with a mean age of 29 years and comprising 109 females and 108 males, available here: https://datadryad.org/dataset/doi:10.5061/dryad.2v6wwpzt3. We also analyzed the data by pooling all the data from the "Florida + London" group, which comprised 551 participants. For more details on the Polish and London datasets, the readers are advised to look at the Jankowska & Karwowski (2022) and Clark & Maguire (2023) publications, respectively. In each population, participants either performed the VVIQ-2 or the PSIQ questionnaires, as instructed. Polish data was obtained from participants doing the Polish translated version of VVIQ-2. In contrast, all other populations performed the imagination surveys in the original English version. We did not have a prior sample size for our main psychological network analysis; rather, we computed CS-coefficient (discussed below) to estimate the reliability of centrality measures. We synthesized the 600 samples of the LLM population based on human vividness scores to estimate an imagination network across different conversation conditions (independent and cumulative) for an imagination instrument, as described below.

**AI Models**

We used six models in total, with four from the Gemma3 family and two models from the Llama family. From the Gemma3 family, we utilized models with 12 and 27 billion parameters, along with their corresponding quantization-aware trained models (postfixed as QAT). Quantized models often reduce model size without changing the number of parameters and quality of responses by lowering the decimal precision of the model weights, enabling us to compare models across different training regimes while keeping the model architecture constant. For more information about the quantization-aware training, readers are recommended to refer to Jacob et al. (2017). From the Llama family, we used Llama-3.3, which has 70 billion parameters, and Llama-4.0-Scout (16x17b), a model that combines an expert's model with 16 experts and 17 billion parameters each. Thus, each model offers a distinct type of long-term memory, based on its training regime and architecture. All models are openly and freely available using Ollama (https://ollama.com/) and were run locally using one NVIDIA B100 GPU.

**Imagination Tasks**

The VVIQ-2 and PSIQ ask participants to rate the vividness of their imagination on a scale of 1 to 5 and 0 to 10, respectively, for a given item prompt. We assume that vividness ratings capture a psychological experience based on the perceived subjective quality of imagined experience. The VVIQ-2 consists of eight contexts with four items each with 32 items in total. We used three items in each of the PSIQ's contexts, with 21 total imagination prompts. We hypothesized that vividness ratings exhibit structural characteristics organized by memory, with covariation patterns replicating across populations. In the current investigation, we utilized the VVIQ-2 and PSIQ to characterize the covariation structure of vividness ratings across imagined scenarios and measure the structural properties of human and LLM imagination using psychological network science. Inherently, the VVIQ-2 and PSIQ are structurally clustered and organized based on different contexts (8 contexts in VVIQ-2 and 7 contexts in PSIQ, as mentioned above). Additionally, the VVIQ-2 contexts focus solely on the visual aspects of the external world, whereas the PSIQ contexts sample the ability to generate and experience sensory imagery across multiple modalities.

For the simulation of LLM responses, the English version of the imagination questionnaires was converted into JSON format and can be found on the OSF. AI model providers frequently use JSON to format text for training the LLM models (He et al., 2024). Using explicit output parsers, our LLMs were constrained to provide only vividness rating responses, and with our LLM-based AI models, we performed the imagination task in two ways: (1) independently and (2) cumulatively. In the independent condition, LLM models responded to each context and prompt without recalling previous trials, treating each item as independent of the others as instructed to the participants. During the cumulative condition, LLM models had access to previous interactions and values; thus, only the first item in this condition did not have information about the past, while subsequent items had a memory of past interactions with items/prompts and their own vividness rating. Thus, our two tasks allowed us to simulate vividness ratings with either (1) independence of previous trials, or (2) having previous trial memory, in LLM models for simulation.

The system message to each LLM call was provided in the following format:

*{“You are a helpful assistant with the following individual characteristics:*

*<definition of imagination ability> <persona statements>*

*Perform the task as per the instructions described below.*

*TASK CONTEXT: <instructions>” }*

An example trial for the VVIQ-2 given to the LLM looked like the following: “{"TRIAL_CONTEXT": "Think of the rising sun. Consider carefully the picture that comes before your mind's eye.", "TRIAL": "A rainbow appears."}". The trial context and message changed depending on the questionnaire. An example trial for the PSIQ looked like: *“Imagine the appearance of a bonfire. How vivid is the image?”* The item from each questionnaire to the LLM was provided as Human/User messages for the LLMs to provide the vividness rating as responses in JSON format: {Vividness: x}, depending on the scale of the given imagination questionnaire.

Each model was used to simulate 200 personas randomly sampled from the Persona-Chat dataset from Zhang et al. (2018), across five levels of imagination ability characteristics. To simulate different imagination abilities, we manipulated them across five levels with prefixed imagination ability personas: (a) no imagination ability prompt; (b) typical imagery: *“Typical imagery is a standard mental imagery capacity most people have. It allows for moderately vivid and detailed mental pictures. Individuals with typical imagery can summon clear, detailed images in their mind's eye but might not be able to sustain them for long periods or with photographic precision”*; (c) aphantasia: *“Aphantasia is inability to form mental images, even* *when attempting to do so. Individuals with aphantasia typically describe a "blank mind's eye" when asked to imagine something, such as a loved one's face or a scene."*; (d) hypophantasia: *"Hypophantasia is reduced ability to form mental images. Visualizations may exist but are vague, dim, or not detailed. Individuals with hypophantasia may perceive faint or blurry images but lack the richness and clarity of typical imagery"*; and (e) hyperphantasia: *"Hyperphantasia is the ability to form extremely vivid, lifelike mental images that are nearly indistinguishable from actual vision. Individuals with hyperphantasia may report "seeing" images as if they were real, with high detail, colors, and sometimes motion."* When no imagination ability prompt was given, only the persona statements were provided with the task instructions, as in the VVIQ-2 or PSIQ questionnaire. For the simulations involving all other imagination ability characteristics, persona statements were prefixed to the definition of the imagination ability. Thus, in total, for a type of task (independent or cumulative), 1000 simulations were generated from each LLM model, with 200 simulations for each of the five imagination ability instruction conditions. The total vividness score from the VVIQ-2 was analyzed using JASP Team (2023) to investigate the effects of LLM task type (cumulative or independent), imagination ability prompt, and model (Fig. 1A). Kruskal-Wallis and Dunn's post-hoc tests were performed for between-group comparisons, and two-tailed p-values are reported in the Results section. We did not have a hypothesis for the total vividness score, other than anticipating that imagination ability prompts for aphantasia would cause total vividness scores to be lower, and hyperphantasia prompts would cause vividness scores to be higher compared to the typical imagery instruction or no imagination ability instruction.

**AI Population Diversity Sampling**

For constructing the imagination networks from each of the LLM responses, we created a population of LLM simulations based on human imagination abilities by downsampling 1000 simulations from across imagination abilities to 600 total simulations for each of the LLMs. The human population consists of individuals with varying imagination abilities, as measured using the total vividness score given a questionnaire. We first extracted the quantile ranges using every 10th percentile of the total vividness scores from the pooled human data from the VVIQ-2 (Florida + Poland group) and PSIQ (Florida + London group). For each LLM simulation, given the model and questionnaire, the total vividness score was computed. LLM simulations were selected based on the human quantile ranges, and from each quantile range, 60 LLM simulations were randomly chosen. Whenever there were not enough simulations for a given quantile range, all simulations were selected within the given quantile range. The remaining simulations for that quantile were chosen from outside the quantile range and removed from the overall samples for further sampling. Thus, after population diversity sampling, we had 600 simulations from each of the models for the given task (independent or cumulative) and imagination questionnaire (VVIQ-2 or PSIQ), which were used to create imagination networks.

Additionally, an important reason to perform AI population diversity sampling is that differences in imagination abilities could explain the topological regularities observed in human imagination networks. Diversifying AI simulations based on imagination ability from human data helps us introduce statistical variances for the LLM-based imagination networks, which can represent the distribution of human imagination across different imagination abilities. Total vividness scores for the VVIQ-2 data were analyzed for different populations from human groups and LLM models after population diversity sampling (Fig. 1B) using Kolmogorov-Smirnov (K-S) tests with Benjamini/Hochberg FDR corrections for multiple comparisons, with two-tailed p-values reported. The code can be found on OSF.

**Network Analysis**

Network analysis of vividness ratings allows us to characterize how imagined scenarios relate to one another within the covariation structure of vividness. We constructed edge weights between two nodes to represent the regularized partial correlations in vividness ratings between pairs of imagined scenarios, yielding an imagination network for each population. Each node represents an imagined scenario, and each edge represents the strength of vividness association between two scenarios, forming what we term “imagination networks" in the current investigation.

Different network measurements represent how information is processed in the networks. Given a specific imagination network, we characterized the representational organization of each human or LLM population using micro-level network centrality of each node, which reveals the importance of each node in a given network. Although there are numerous centrality measures in network science, we focused on four common psychological network centrality measures: expected influence, strength, closeness, and betweenness (Borsboom et al., 2021), using the bootnet library (Epskamp & Fried, 2025) in R. In our study, we correlated the different centralities of nodes of an imagination network with those from all other networks to investigate whether each imagined scenario held similar structural importance across the various networks for a given imagination questionnaire. If a centrality measure for the nodes in an imagined network correlated with another imagined network, we inferred that the imagined scenario held similar importance in the covariation structure across populations. Thus, a high positive correlation of centrality measures across imagination networks indicates that the covariation structure of vividness is organized similarly across populations.

We also performed post-hoc stability analysis; specifically, we computed the CS-coefficient using the bootnet library with a case-dropping value of 500 and a bootstrap value of 5000 for the centrality measurements to establish their robustness and accuracy. Centrality stability is preferred to be above 0.5, or at least above 0.25 (Epskamp et al., 2018; Epskamp & Fried, 2018). The CS-coefficient (centrality stability) effectively measures the proportion of data that can be eliminated while ensuring, with 95% confidence, that a correlation of at least 0.7 is maintained with the original centrality measures (Epskamp et al., 2018; Epskamp & Fried, 2018). See Supplementary Table S1 for VVIQ-2 and Table S2 for PSIQ for CS-coefficient results. We investigated whether positive centrality correlations would emerge across populations, indicating shared representational organization, and report pairwise Pearson's r correlations and one-tailed p-values. Whenever pairwise normality was violated, we provided Spearman's rho correlations. Further, we also reported Benjamini/Hochberg FDR corrections for multiple comparisons for the p-values.

The vividness ratings are ordinal scale data for which polychoric partial correlations (Epskamp et al., 2018; Epskamp & Fried, 2018) should be preferred, as they assume a latent distribution underlying the ordinal scale value. However, our initial investigations failed to find stable centrality measures in networks, primarily for the LLMs. Thus, the Spearman method was used for partial correlations in EBICglasso for estimating the imagination networks (Epskamp et al., 2018; Epskamp & Fried, 2018). To investigate whether nodes had similar influences across networks, depending on the type of centrality measures, we performed pairwise correlations between the centralities of nodes across networks.

To understand how different items are in the data clusters based on the associations of vividness ratings between them, we performed clustering analysis using the walktrap community (Pons & Latapy, 2005) detection algorithm with the LE method in the EGANet library (Golino & Christensen, 2025) in R. The results presented here are robust to the type of community detection algorithm as in our preliminary analysis we utilized different algorithms (TMFG, Massara et al., 2016) and found similar results (Ranjan & Odegaard, 2025). The similarity in cluster assignments of different nodes across the networks was calculated using the adjusted rand score from the sklearn library (Buitinck et al., 2013) in Python. The higher the adjusted rand-score, the higher the clustering alignment across networks. We correlated the centralities across populations and computed adjusted rand score to measure the alignment of networks across populations.

Each network was computed using the bootnet library (Epskamp & Fried, 2025) in the R programming language based on the EBICglasso method with a commonly used common tuning parameter of 0.5 using the vividness ratings for each item. The EBICglasso method promotes fewer connections between network nodes by combining Graphical LASSO and the Extended Bayesian Information Criterion for model selection (EBICglasso), incorporating partial correlations, which results in fewer false-positive edges in the network compared to networks formed solely using partial correlations . See Table S1, S2 for clustering assignment for each node for VVIQ-2, and PSIQ imagination networks respectively. The code can be found on OSF.

## Results

### Human and LLM populations show distinct vividness distributions

We used vividness ratings from human participants across two validated imagination questionnaires: the VVIQ-2 (32 items spanning 8 environmental scene contexts with a 1 to 5 point scale) from Florida (N = 541) and Poland (N = 1,651), and the PSIQ (21 items across 7 sensory modalities with a 0 to 10 point scale) from Florida (N = 334), London (N = 217), and Florida+London (N = 551). For the VVIQ-2, we randomly subsampled 2 non-overlapping groups of N = 600 from the Poland group (N = 1,651) to compare the total vividness, centrality and clustering within groups was well as across groups with Florida (N = 541) and Florida+Poland (N = 2192). The Florida and London groups performed PSIQ and VVIQ-2 in English and the Polish group performed the VVIQ-2 in the Polish language. To enable comparison with humans, we generated LLM responses using six model variants: Gemma3-12B, Gemma3-27B, their quantization-aware trained counterparts (Gemma3-12B-QAT, Gemma3-27B-QAT), Llama3.3-70B, and Llama4x16x17B (mixture-of-experts architecture) by matching the total vividness distribution with the human groups *(See Methods)*. Each model produced 1,000 simulations by crossing 200 synthetic personas (Zhang et al., 2018) with five imagination ability conditions: *aphantasia* (no imagery), *hypophantasia* (reduced imagery), t*ypical imagery*, *hyperphantasia* (extremely vivid imagery), and *no explicit instruction.* Their effects can be observed in *Fig. 1A*. We tested two conversational conditions: independent (stateless, with each item rated without any memory of previous responses) and cumulative (with conversational memory maintained across sequential items).

Total vividness scores (the sum of scores across all the questionnaire items) revealed systematic differences between LLMs with the imagination condition ability prompt. Specifically, for the VVIQ-2 (theoretical range 32-160), Kruskal-Wallis tests demonstrated significant main effects of imagination ability instruction ($H(4)=9{,}588$, $p<10^{-7}$), conversational condition ($H(1)=49.2$, $p=2\times10^{-12}$), and model architecture ($H(5)=461$, $p<10^{-97}$) as shown in *Fig. 1A*. Dunn's post-hoc tests revealed that scores increased monotonically with prompted imagination ability, from aphantasia instructions (mean=43.2, 95% CI [42.8, 43.6]) to hyperphantasia (mean=141.3, 95% CI [140.7, 141.9]). Unexpectedly, the no-instruction condition produced higher scores (mean=114.6, 95% CI [114.0, 115.2]) than explicit typical imagery prompts (mean=107.8 [107.2, 108.4]), suggesting LLMs default to vivid imagination absent explicit constraints. The cumulative conversational condition increased total scores by 5.4% relative to independent (97.4 vs 92.3), and scores scaled systematically with model size (Gemma3-12B: 88.3, 95%CI [87.7, 88.9]; Llama4-16x17B: 107.1, 95%CI [106.4, 107.8]). Similar patterns emerged for the PSIQ (Fig. S1A).

To construct LLM populations matching human imagination ability diversity, we downsampled each model's 1,000 simulations to 600 based on human total vividness score distributions (see Methods). Kolmogorov-Smirnov tests revealed that all LLM distributions differed significantly from all human populations (all $D>0.15$, $p<0.001$ after FDR correction; Fig. 1B). Within humans, most populations showed statistically similar distributions (exceptions included Florida vs Florida+Poland $D=0.07$, $p_{BH}=0.035$; Poland-1 vs Poland-All $D=0.11$, $p_{BH}<0.001$), confirming that human imagination ability profiles are conserved across geographic populations (total vividness scores and their distribution are displayed in *Fig. 1B*).

These univariate analyses establish that LLMs can produce controllable vividness responses sensitive to prompting manipulations. However, total vividness scores, being simple sums across items, provide no insight into how imagined scenarios relate to one another within a cognitive agent's internal representations. A person or model could achieve identical total scores through entirely different patterns of scenario-specific vividness, reflecting fundamentally different organizational structures. To characterize the relational structure of vividness, we therefore turned to multivariate network analysis, which captures how vividness patterns across scenarios reveal the underlying representational organization that univariate statistics cannot detect.

A. LLM's mean vividness across imagination ability.

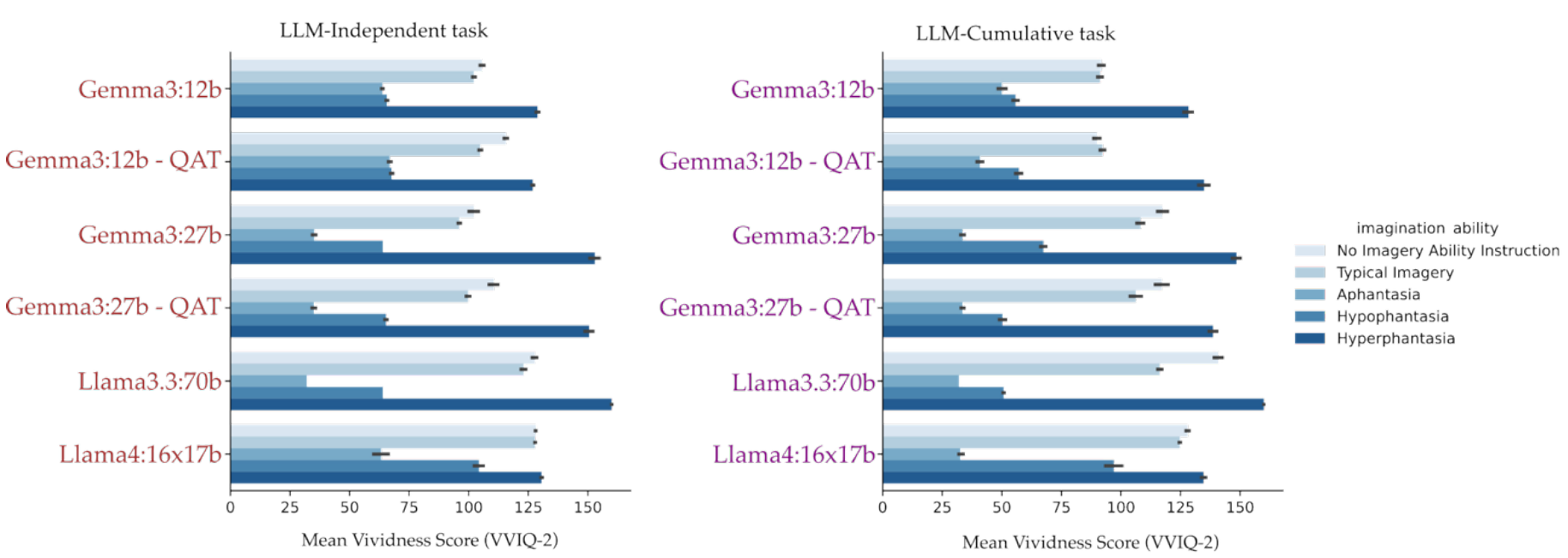

B. Mean (left) and distribution (right) of total vividness score in Humans and after population diversity sampling in LLMs.

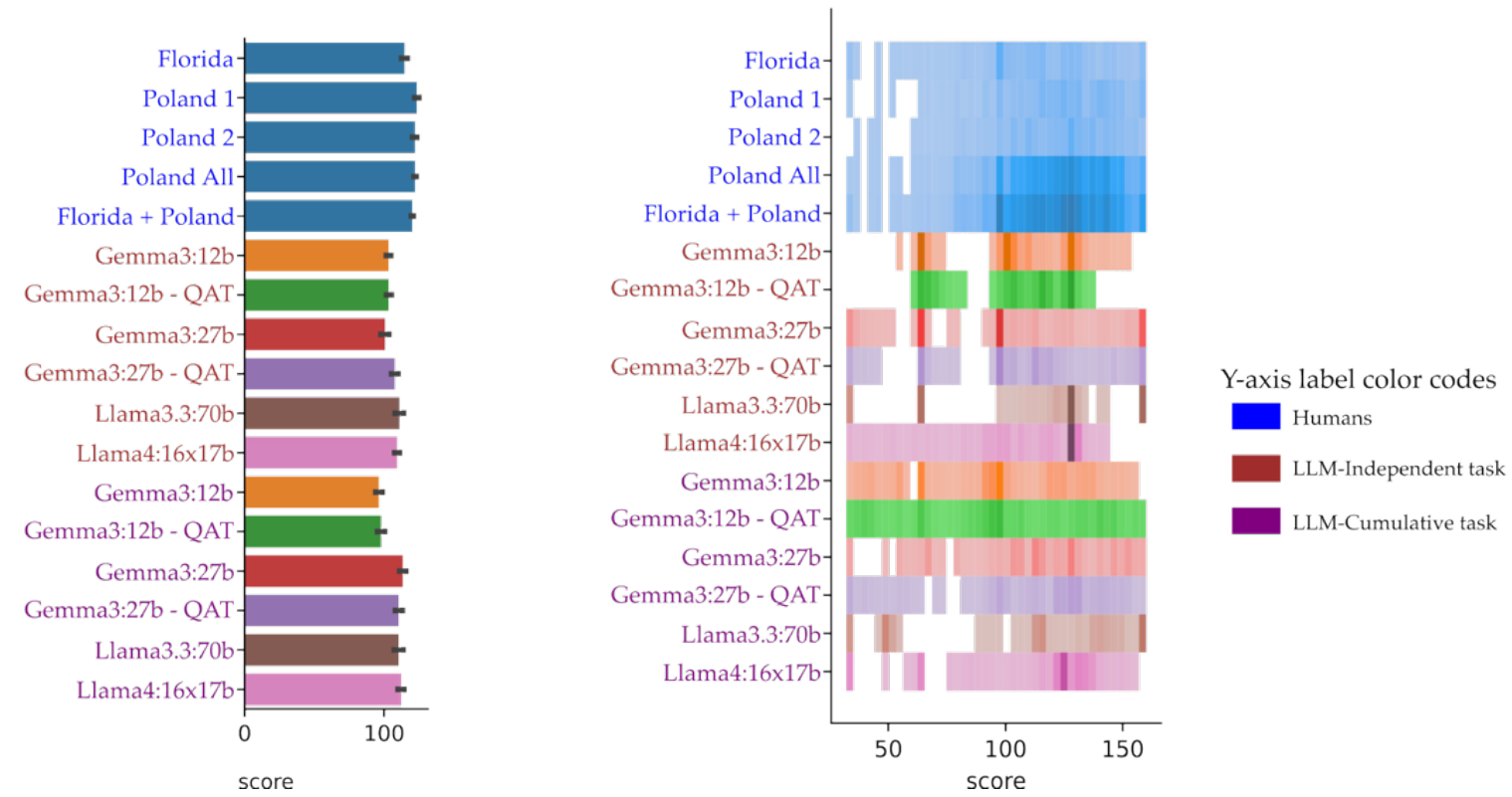

**Fig 1. Total VVIQ-2 vividness scores and distributions in LLMs and human populations.** (A) Mean vividness scores across different VVIQ-2 LLM simulations. The total score is the sum of all vividness ratings in the questionnaire in independent item sampling (left) and conversational memory sampling (right). For each LLM in a task (independent or cumulative) with a total 1000 simulations generated using 200 personalities and five imagination ability conditions. (B) (*Left*) Average total vividness scores from human populations in (blue) Florida ($N = 541$), Poland-1 ($N = 600$), Poland-2 ($N = 600$), Poland-All ($N = 1651$), Florida+Poland ($N = 2192$). Each LLM agent after population diversity sampling (to match variability in the imagination abilities of LLMs with humans; see Methods), with 600 simulations from different imagination abilities. (*Right*) Distributions of the total vividness scores in humans and LLM agents for the VVIQ-2. Darker colors denote a higher density of data in that bin for a given population of cognitive agents (human or LLM). See *Fig. S1* for vividness scores for the PSIQ. The error bars in bar plots are bootstrapped ($N = 1000$) *95%* confidence intervals.

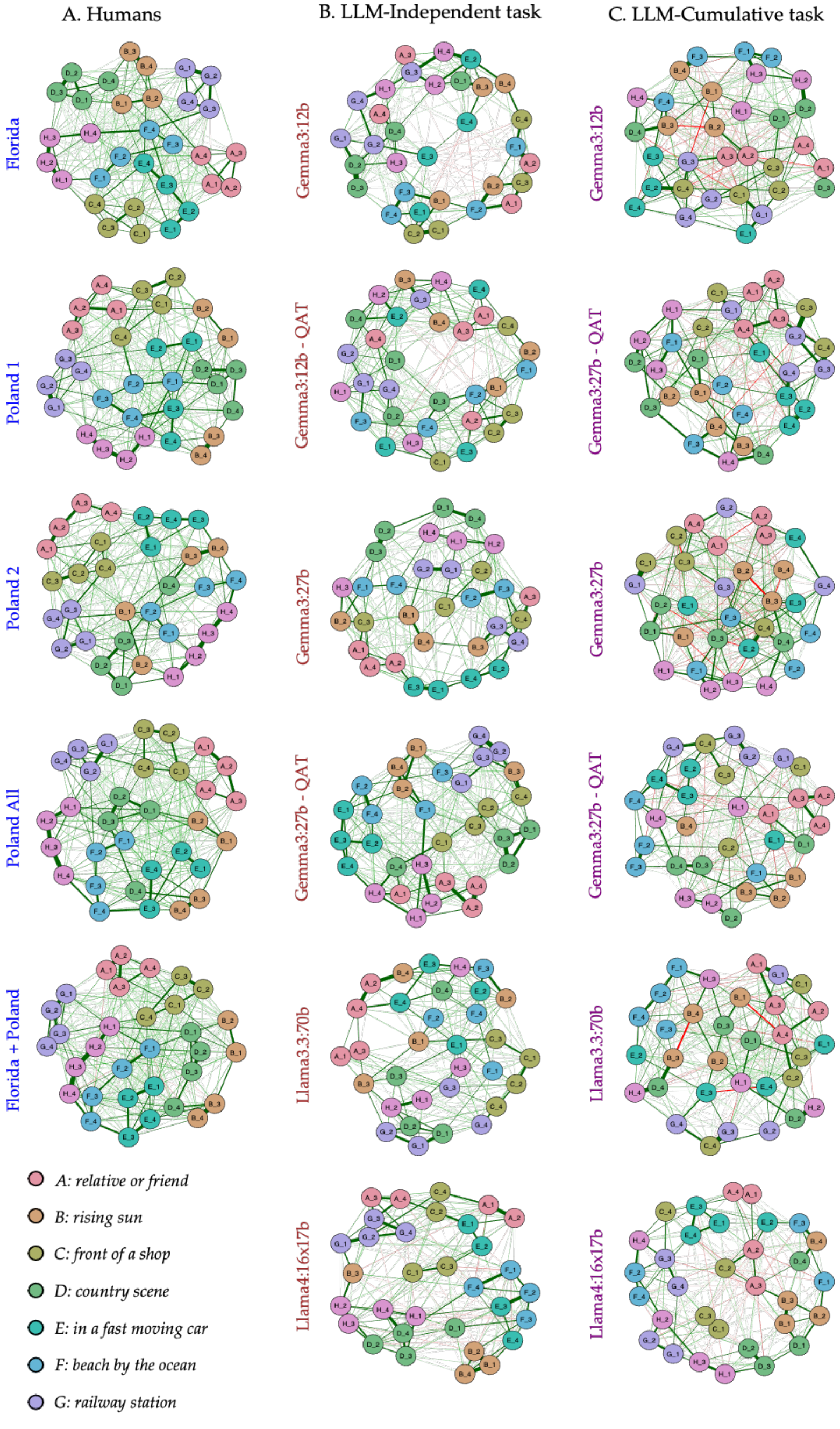

**Fig. 2. Imagination networks using the VVIQ-2 in Humans and LLMs.** Cognitive agents (humans or LLMs) report the vividness of imagined items across different contexts of the environment, denoted as different colors in the networks. There are four items in each context, with postfixed numerals labeling each node. The vividness rating is reported on a scale of 1 to 5. The network was estimated using the EbicGlasso method using Spearman partial correlations of vividness ratings between different pairs of nodes. (A) Imagination networks in human populations: Florida ($N = 541$), Poland-1 ($N = 600$), Poland-2 ($N = 600$), Poland-All ($N = 1651$), Florida+Poland ($N = 2192$), from top to bottom. (B) LLM imagination networks from Gemma and Llama variants in independent task conditions. (C) In the LLM imagination networks in the cumulative task condition, each scene's vividness rating is provided utilizing the conversation history of previous interactions (i.e., previous responses to items on the questionnaire). Each LLM network consists of N = 600 simulations. The color of nodes signifies the context in the VVIQ-2 with four items in each context. The color and thickness of an edge indicate its magnitude and direction of association. Red lines indicate negative associations, and green lines indicate positive associations for an edge.

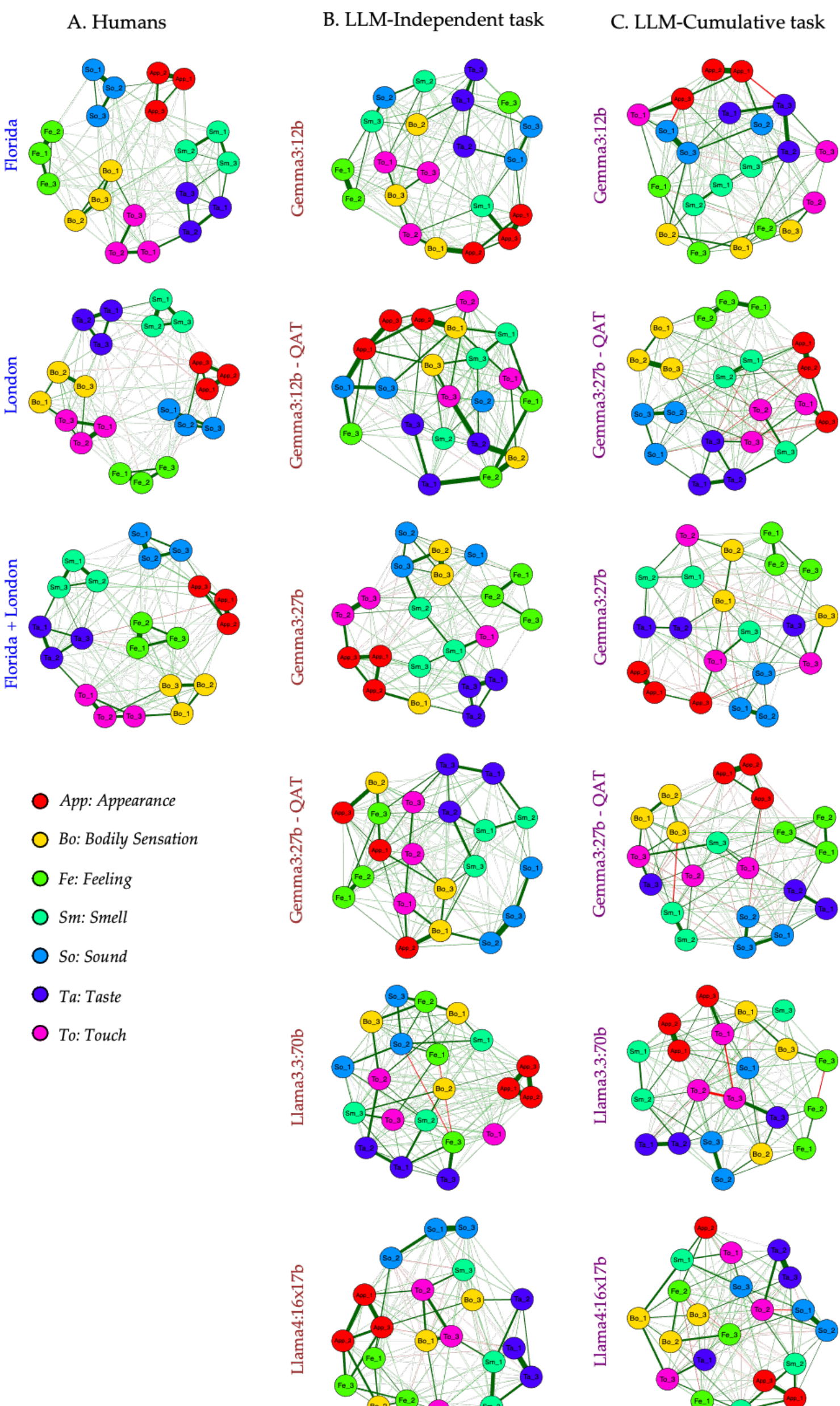

**Fig. 3. Imagination networks using the PSIQ in Humans and LLMs.** Cognitive agents (humans or LLMs) report the vividness of imagined items across different contexts of sensory experiences, denoted as different colors in the networks. There are three items in each context, with postfixed numerals to each node. The vividness rating is reported on a scale of 0 to 10. The network is estimated using the EbicGlasso method using Spearman partial correlations of vividness ratings between different pairs of nodes. (A) Imagination networks in human populations: Florida (N = 334), London (N = 217), and Florida+London (N = 551), from top to bottom. (B) LLM imagination networks in the independent task condition. Each node's vividness rating is rated independently. (C) In the LLM imagination networks in the cumulative task condition, each item's vividness rating is rated with the conversation history of previous interactions. Each LLM network consists of N = 600 simulations. The color of nodes signifies the modality in the PSIQ, with three items in each modality. The color and thickness of an edge indicate its magnitude and direction of association. Red lines indicate negative associations, and green lines indicate positive association for an edge.

**Expected influence and strength reveal high correlations within human populations, highlighting the importance of nodes in imagination networks**

We constructed imagination networks from vividness rating matrices using regularized partial correlations (EBICglasso method with Spearman correlations (Borsboom et al., 2021; Epskamp et al., 2018; Epskamp & Fried, 2018; Figs. 2, 3). Local centrality measures included expected influence and strength, which quantify node importance through direct network connections; these measures revealed striking structural consistency across human populations.

For the VVIQ-2 expected influence centrality, within-Poland population centrality correlations were strong: Poland-1 vs Poland-2 ($r=0.712$, 95% CI [0.525, 0.833], $p<10^{-5}$), Poland-1 vs Poland-All ($r=0.931$ [0.882, 0.961], $p<10^{-15}$), and Poland-2 vs Poland-All ($r=0.860$, $p<10^{-10}$). Cross-population comparisons between Florida and Poland groups showed moderate to strong correlations: Florida vs Poland-1 ($r=0.589$ [0.347, 0.754], $p=2\times10^{-4}$, $p_{BH}=0.002$), Florida vs Poland-2 ($r=0.466$, $p=0.0036$, $p_{BH}=0.030$), and Florida vs Poland-All ($r=0.595$, $p=1.6\times10^{-4}$, $p_{BH}=0.002$). Critically, composite populations containing overlapping participants showed enhanced correlations compared to constituent groups: Florida vs Florida+Poland ($r=0.732$ [0.554, 0.845], $p<10^{-6}$), Poland-1 vs Florida+Poland ($r=0.920$, $p<10^{-15}$), and Poland-All vs Florida+Poland ($r=0.999$, $p<10^{-30}$; Fig. 4A-1). Similar patterns emerged for the human populations' PSIQ imagination networks: Florida vs London ($r=0.425$, $p=0.027$), Florida vs Florida+London ($r=0.914$ [0.810, 0.963], $p<10^{-9}$), and London vs Florida+London ($r=0.733$, $p=7.9\times10^{-5}$; Fig. 4B-1).

Strength centrality, the sum of unsigned edge weights, showed similar results across human populations. For the VVIQ-2: Florida vs Poland-1 ($r=0.591$, $p=1.8\times10^{-4}$, $p_{BH}=0.003$), Poland-1 vs Poland-2 ($r=0.692$, $p<10^{-5}$), Poland-1 vs Poland-All ($r=0.928$, $p<10^{-15}$), and Florida vs Florida+Poland ($r=0.725$, $p<10^{-6}$; Fig. 4A-2). For PSIQ: Florida vs Florida+London ($r=0.858$, $p<10^{-6}$), and London vs Florida+London ($r=0.538$, $p=0.006$; Fig. 4B-2).

**In stark contrast, LLM networks showed weak and inconsistent correlations with human networks.**

Of the possible LLM-human pairwise comparisons for VVIQ-2 expected influence, only 8 achieved significance after FDR correction: Gemma3-27B-QAT in the independent condition showed weak correlations with human populations ($r=0.308$-$0.393$, $p_{BH}=0.04$-$0.08$), while four models in the cumulative condition showed modest improvements (Gemma3-12B, Gemma3-12B-QAT, Gemma3-27B, Llama3.3-70B; maximum $r=0.548$ for Gemma3-12B-QAT vs Poland-All, $p_{BH}=0.007$). The PSIQ showed only 6 of 72 possible comparisons reaching significance. Median LLM-human expected influence correlation was $r=0.13$ for VVIQ-2 and $r=0.09$ for PSIQ, compared to median human-human correlations of $r=0.66$ (VVIQ-2) and $r=0.76$ (PSIQ), representing an approximately 5-fold difference in structural alignment.

For strength centrality in the VVIQ-2, only Gemma3-12B and Gemma3-27B-QAT showed significant correlations in independent condition ($r=0.302$-$0.452$ across human populations), while the cumulative condition yielded sporadic significance that rarely survived multiple comparison correction. PSIQ strength showed similar patterns: Gemma3-12B correlated with all human groups in independent mode ($r=0.404$-$0.422$), but most other configurations showed null results. Overall, local centrality measures demonstrate that human populations share highly conserved network organizational principles that current LLMs fail to replicate.

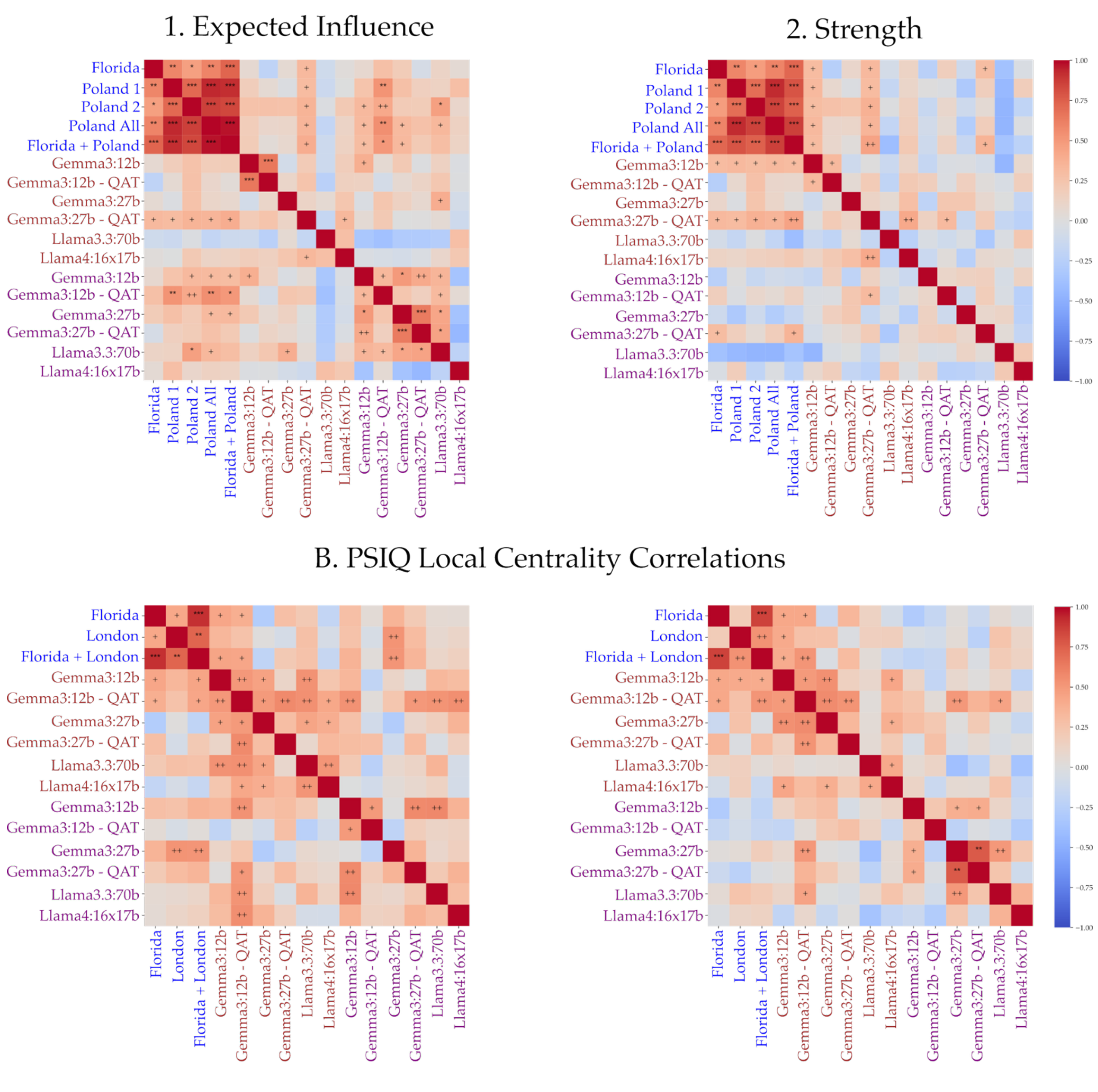

**Fig. 4. Pearson's r correlation heatmaps of node importance using different local centrality measures across VVIQ-2 and PSIQ imagination networks**. The left column denotes correlations for expected influence; the right column denotes correlations for strength. (A) Centrality correlations among cognitive agents for the nodes in VVIQ-2 imagination networks with 32 nodes (*top row*). (B) Centrality correlations among cognitive agents for the nodes in PSIQ imagination networks with 21 nodes (*bottom row*). Expected influence and strength centralities exhibited high correlations among human populations for the VVIQ-2, and a similar pattern of results was found for the PSIQ. + denotes a significant result based on uncorrected p-values and * denotes significant result based on corrected p-values using Benjamini/Hochberg FDR correction for multiple comparisons; */+ $< 0.05$; **/++ $< 0.01$; *** $< 0.001$. The heatmaps are symmetric across the top-left to bottom-right diagonal.

**In humans, imagined nodes are highly correlated for closeness, but not for betweenness centrality**

Closeness centrality is the measurement of how efficiently information flows from a node to all other nodes via the shortest network paths; this measure revealed robust correlations across human imagination networks. For the VVIQ-2, all human population pairs showed significant positive correlations: Florida vs Poland-1 ($r=0.535$, $p=7.9\times10^{-4}$, $p_{BH}=0.014$), Florida vs Poland-2 ($r=0.313$, $p=0.041$), Florida vs Poland-All ($r=0.572$, $p=3.1\times10^{-4}$, $p_{BH}=0.006$), Florida vs Florida+Poland ($r=0.776$, $p<10^{-6}$), Poland-1 vs Poland-All ($r=0.619$, $p=8.0\times10^{-5}$, $p_{BH}=0.002$), Poland-2 vs Poland-All ($r=0.640$, $p=4.0\times10^{-5}$, $p_{BH}=0.001$), and Poland-All vs Florida+Poland ($r=0.884$, $p<10^{-10}$; Fig. 5A-1). PSIQ networks exhibited even stronger closeness correlations: Florida vs London ($r=0.759$, $p=3.3\times10^{-5}$, $p_{BH}=0.001$), Florida vs Florida+London ($r=0.874$, $p<10^{-7}$), and London vs Florida+London ($r=0.842$, $p<10^{-6}$; Fig. 5B-1).

LLM networks showed inconsistent closeness correlations with humans, though with notably better performance than local centrality measures. For the VVIQ-2, correlations remained sparse (9 of 144 comparisons significant), but the PSIQ showed modest improvement (15 of 72 comparisons significant), suggesting LLMs may better approximate sensory concept organization than environmental scene structure but still lower than that of between human population groups comparisons.

Betweenness centrality quantifies how frequently a node appears in shortest paths between other node pairs; this measure showed universally weak correlations across all cognitive agent comparisons (median $r=0.08$ for all pairwise comparisons; Fig. 5A-2, 5B-2). Furthermore, betweenness demonstrated poor stability in case-dropping bootstrap analysis (CS-coefficient<0.25 for most networks; Tables S1-S2), indicating high sampling variability. The weak correlations and low stability suggest that betweenness captures individual differences in how scenarios bridge different regions of imagination networks, likely reflecting personalized experiential histories that vary substantially even within populations sharing similar overall network structure.

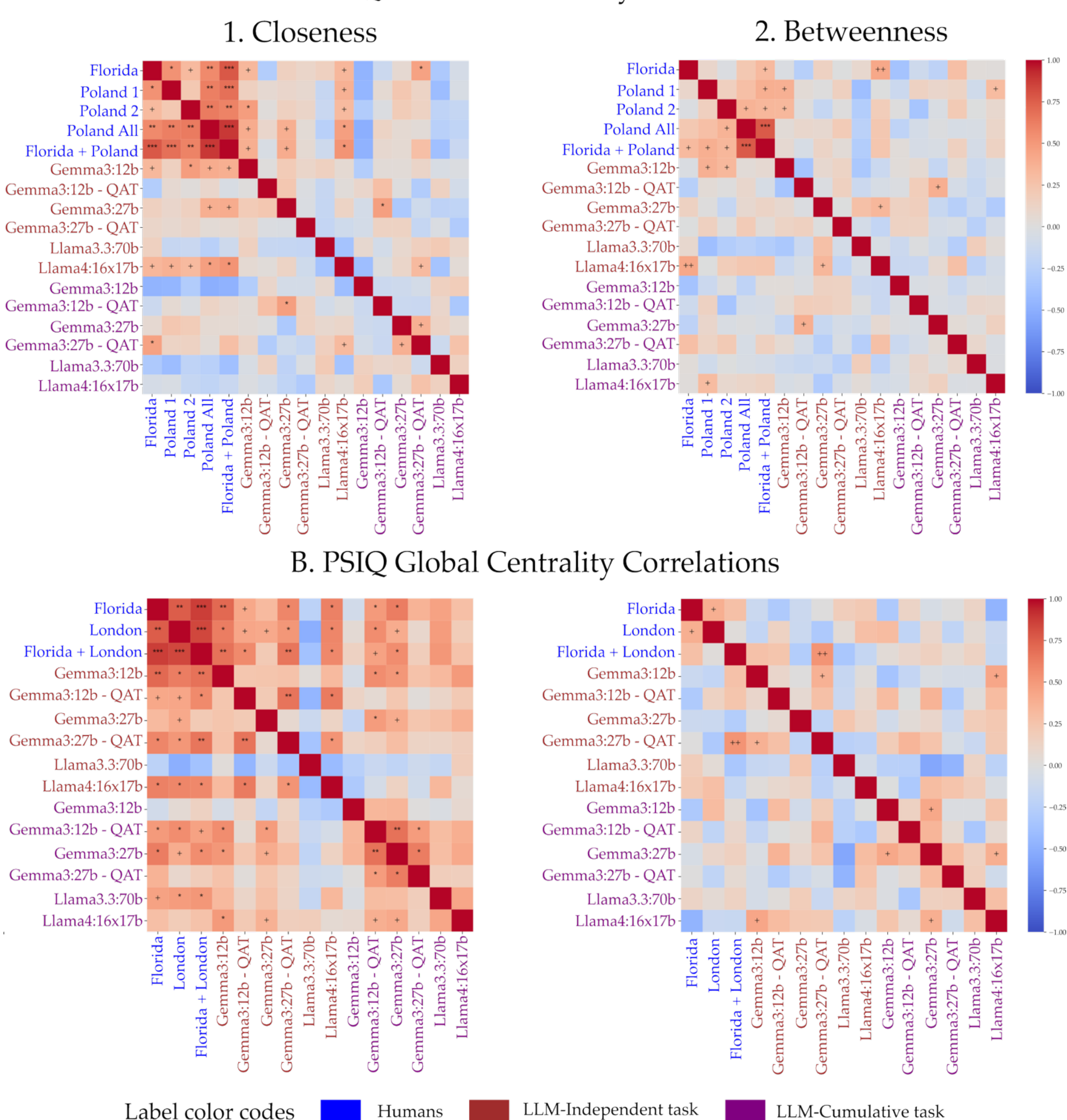

**Fig. 5. Pearson's r correlation heatmaps of node importance using different global centrality measures across VVIQ-2 and PSIQ imagination networks.** Columns, left to right, display different centrality measures; specifically, closeness and betweenness. (A) Centrality correlations among cognitive agents for the nodes in VVIQ-2 imagination networks with 32 nodes (*top row*). (B) Centrality correlations among humans but not in betweenness for the nodes in PSIQ imagination networks with 21 nodes (*bottom row*). Closeness centrality correlations across cognitive agents show high correlations among human populations across VVIQ-2 and PSIQ. Betweenness centrality results show a lack of centrality correlations across cognitive agents in both VVIQ-2 and PSIQ. + denotes significance based on uncorrected p-values and * denotes significance based on corrected p-values using Benjamini/Hochberg FDR correction for multiple comparisons; */+ < 0.05; **/++ < 0.01; *** < 0.001. The heatmaps are symmetric across the top-left to bottom-right diagonal.

## Clustering is prominent in human networks and often absent in LLMs for imagination networks.

Data-based clusters were computed from the imagination networks (Fig. 2 and Fig. 3) and calculated using the walk-trap algorithm (Pons & Latapy, 2005) to measure the meso-level properties of the imagination networks. As centralities capture the micro-level properties of the network at the node level, one can investigate the meso-level characteristics of the network in terms of how a group of nodes cluster together (Bringmann et al., 2019; Fortunato, 2010). The walk-trap algorithm is a community detection algorithm that identifies communities/clusters of nodes characterized by dense connections among themselves, rather than to other nodes in the network. Community detection (walk-trap algorithm) revealed qualitative differences in network organization. Human VVIQ-2 networks decomposed into 4-6 modules reflecting questionnaire structure (Table S3), while 5 of 6 LLM models in independent conditions produced single-cluster networks. Cumulative memory increased clustering for some models (Gemma3-27B-QAT: 5 clusters), but patterns remained model-dependent. The PSIQ showed sharper contrasts: human networks yielded 4-5 clusters versus predominantly single-cluster LLM networks, except Gemma3-27B-QAT in cumulative mode (5 clusters; Table S4).

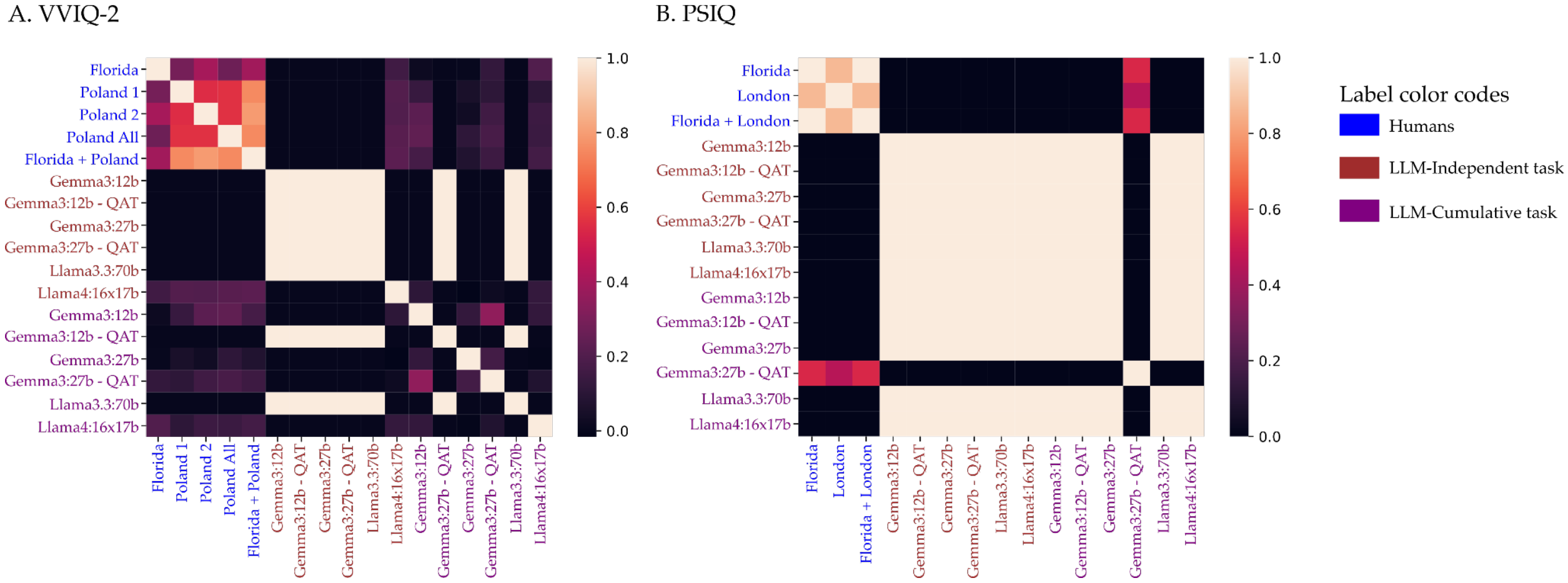

**Fig. 6. Alignment of clusters across imagination networks as measured using the Adjusted Rand Index.** (A) Clustering alignment in the imagination network from the VVIQ-2 task. (B) Clustering alignment in the imagination network from the PSIQ task. The heatmap matrices are symmetric, from the top-left to the bottom-right diagonal.

In Fig. 6*,* the adjusted Rand Index (ARI) quantified clustering similarity (1.0=perfect alignment, 0=chance, negative=worse than chance). Human VVIQ-2 networks showed moderate alignment (ARI=0.27-0.40), while PSIQ networks exhibited strong alignment (ARI=0.87-1.0; Fig. 6). Most LLM independent-task networks achieved ARI=0 with humans due to degenerate single-cluster topology. The highest-performing model (Llama4-16x17B independent) reached ARI=0.22 (VVIQ-2). Cumulative conversation memory improved some models (Gemma3-27B: ARI=0.30 for VVIQ-2; Gemma3-27B-QAT: ARI=0.54 for PSIQ), yet all remained below human-human alignment.

## Discussion

Vividness is an important psychological descriptor of the strength of mental experience (Fazekas, 2024; Morales, 2023) which can be communicated linguistically (Langkau, 2021). Using imagined scenes and the association of vividness ratings between them, we constructed psychological imagination networks across sensory scales using commonly used instruments (VVIQ-2 and PSIQ) for measuring imagination ability. The centrality and clustering alignment observed across human imagination networks, consistent across the VVIQ-2 and the PSIQ and across populations from Florida, Poland, and London, was expected given that vividness covariation should reflect shared memory organization accumulated through common environmental and sensory experience (Barsalou, 2008; Hassabis & Maguire, 2009; Schacter et al., 2008; Schacter & Addis, 2007a). The near-perfect clustering conservation in the PSIQ (ARI = 0.87-1.0) compared to the moderate conservation in the VVIQ-2 (ARI = 0.27-0.40) was also anticipated: sensory modality boundaries are neurobiologically determined and broadly invariant across individuals and cultures (Barsalou, 2008; Bartolomeo et al., 2026, 2026; Mishkin et al., 1983; Moscovitch et al., 2016; Spagna et al., 2024a; Tulving, 1983), whereas environmental scene contexts rest on episodic memory traces shaped by culturally specific experience (Liang et al., 2024; Schacter & Addis, 2007a), making them less tightly conserved. That both instruments showed centrality and clustering alignment indicates the structural regularity of imagination networks generalizes across sensory scales, from purely visual imagery (VVIQ-2) to seven-modality sensory (PSIQ) imagination. This is possible as the long-term memory of experiences having different features are stored in a dependent manner (Balaban et al., 2020) resulting in an existence of a world model in an agent due to experiences' utility (Land, 2014; Richens et al., 2025).

By contrast, the complete failure of all six LLMs to reproduce human network structure was unexpected in light of prior work showing that LLMs can approximate human perceptual similarity judgments for color and other sensory dimensions (Kawakita et al., 2023; Marjieh et al., 2023; Marjieh, Kumar, et al., 2024; Wang et al., 2025). The key difference is that perceptual similarity ratings can be inferred from co-occurrence statistics in training text, whereas vividness network structure reflects how memory substrates for imagined scenarios overlap within an embodied cognitive system such as humans, a property that linguistic training alone does not confer (Mahowald et al., 2023; Pezzulo et al., 2024). This failure persisted across independent and cumulative conversational conditions, confirming it is not an artifact of how items were presented.

Balaban & Ullman (2025) propose a physics-graphics framework of mental simulation that offers a principled account of these results. In their framework, mental simulation separates physical simulation (an amodal representation of scene geometry; here, a language agent is a physical simulation of human linguistic patterns) from graphical rendering (sensory experiences, here, vividness), which converts that representation into the phenomenologically detailed image that vividness ratings target. The imagination network is therefore a population-level map of rendering organization, and its cross-population replication implies that rendering draws on memory resources structured similarly across populations with shared experiential histories providing a geometrical link to physics and graphics. Within this account, expected influence, strength, and closeness each capture how richly a scenario integrates into the shared rendering architecture and replicate accordingly, whereas betweenness depends on individual associative pathways that vary within and across populations and does not replicate (median $r = 0.08$; CS-coefficient $< 0.25$ in most networks; Epskamp et al., 2018; Epskamp & Fried, 2018).

One result sits outside this account: closeness centrality in the PSIQ showed stronger LLM-human alignment than any other measure (15 of 72 comparisons significant), though still well below human-human levels. Two explanations are possible. The alignment may be a spurious effect of the small number of comparisons surviving FDR correction, in which case it carries no theoretical weight. Alternatively, global sensory integration structure may be partially recoverable from linguistic co-occurrence when semantically proximate sensory concepts (Kawakita et al., 2024; Marjieh, Sucholutsky, et al., 2024; Wang et al., 2025) in text are also proximate in human imagination networks, suggesting LLMs may have acquired some approximation of human sensory representational proximity through language statistics alone. These possibilities lead to different predictions about multimodal models and can be tested directly in future work.

Several limitations bear on the scope of these conclusions. All three human samples are Western, educated, and predominantly young, so the cross-population replication is not yet evidence of a universal organizing principle. Samples whose lived environments differ structurally, such as populations with distinct food environments for gustatory PSIQ items or different transportation landscapes for the railway station context, are needed to test that claim more rigorously. Although we cover different continents in VVIQ-2 which utilizes English-speaking USA (Florida) and Polish-speaking (Poland) data while PSIQ utilizes only English-speaking USA (Florida) and UK (London) populations. Vividness ratings are also self-report only; objective measures such as binocular rivalry suppression or fMRI decoding (Pearson, 2019) are needed to confirm that network structure reflects phenomenological organization rather than differences in scale use or metacognitive accuracy. Population-level networks also mask individual architecture, and given well-established individual differences including aphantasia and hyperphantasia (Balaban et al., 2023; Liu, 2026; Nanay, 2025), idiographic network analysis (Bringmann, 2021) is a natural next step; recent work further suggests that within-subject designs could map geometrical relationships in imagination, since topological relationships between stimuli influence memory recall and perception.

On the LLM side, synthetic populations were constructed by mixing simulations across five imagination ability conditions to match human total vividness distributions, ruling out the possibility that observed network differences are driven by overall imagination ability differences rather than representational organization. However, all tested models are from the Gemma3 and Llama families; frontier models including GPT-4o, Claude, and Gemini are absent, and some of those incorporate multimodal training that may provide richer sensorimotor grounding. Whether the failure extends to multimodal architectures is a question the present study cannot answer, and testing it directly is the most pressing next step for this line of research as the focus of the current study was to qualitative understand how different model sizes (12B-270B) across conversational memory (independent and cumulative task conditions for LLMs).

These findings carry implications for two areas. For cognitive science, the cross-population replication of imagination network structure gives empirical support to the claim that graphical rendering, as distinct from physical simulation, has measurable population-level regularities in the human cognitive world model (Balaban & Ullman, 2025; Land, 2014) , binding the human physics engine to a graphics rendering engine - a coupling which does not occur in purely language-simulation engines like LLM agents. Network analysis of vividness ratings is a tractable method for characterizing those regularities and comparing them across populations, cultures, and individual difference groups. For AI research, imagination network analysis is a more sensitive benchmark for evaluating generative world models than total-score comparison: a model can match human vividness averages while failing to reproduce the relational structure those averages obscure, and the present data establish two concrete thresholds, a minimum cross-population centrality correlation of $r = 0.31$ and a minimum VVIQ-2 clustering alignment of ARI = 0.27, against which future architectures can be evaluated. Thus, what stands out in this work is that producing plausible vividness language and organizing it the way human memory does are separable capacities, which leads to a practical criterion: the relevant test for an artificial world model is not whether it describes imagined scenarios convincingly, but whether the relational structure of those descriptions mirrors the embodied memory organization that imagination network analysis can measure.

## Conclusion

Human imagination networks show structural regularities that replicate across populations and sensory scales; LLMs, regardless of architecture or scale, do not. The failure is specific to phenomenologically-grounded knowledge organization rather than semantic representation in general, as LLMs succeed on sensory similarity tasks where human memory structure can be approximated from language statistics. At the broadest level, these results support the view that the rendering component of human mental simulation is organized by embodied experience in ways that current language models have not acquired. Future work with multimodal architectures, tested against the benchmarks this study establishes, will determine whether richer sensorimotor grounding is sufficient to close that gap.

**Declaration of generative AI and AI-assisted technologies in the manuscript preparation process.**

During the preparation of this work, the authors used Claude Sonnet 4.6 by Anthropic to edit and refine the manuscript's language and clarity. Generative AI was not used for the research conceptualization, study design, or data analysis. After using this tool/service, the authors reviewed and edited the content as needed and take full responsibility for the content of the published article.